\documentclass{article}

\PassOptionsToPackage{numbers, compress}{natbib}
\usepackage[preprint]{neurips_2024}

\usepackage[utf8]{inputenc} % allow utf-8 input
\usepackage[T1]{fontenc}    % use 8-bit T1 fonts
\usepackage[breaklinks,bookmarksdepth=3,bookmarksopen=true,bookmarksopenlevel=2,bookmarksnumbered=true]{hyperref}
\usepackage{url}            % simple URL typesetting
\usepackage{booktabs}       % professional-quality tables
\usepackage{amsfonts}       % blackboard math symbols
\usepackage{nicefrac}       % compact symbols for 1/2, etc.
\usepackage{microtype}      % microtypography
\usepackage{xcolor}         % colors
\usepackage{enumitem}
\usepackage{amsmath} 
\usepackage{graphicx}
\usepackage{wrapfig}
\usepackage{booktabs}
\usepackage{caption}
\usepackage{array}
\usepackage{wrapfig}
\usepackage{subcaption}

\title{Compositional 3D-aware Video Generation \\ with LLM Director}

\author{Hanxin Zhu\textsuperscript{\rm 1}\thanks{This work is accomplished in Microsoft, April 2024.}, Tianyu He\textsuperscript{\rm 2}, Anni Tang\textsuperscript{\rm 3}, Junliang Guo\textsuperscript{\rm 2}, Zhibo Chen\textsuperscript{\rm 1}, Jiang Bian\textsuperscript{\rm 2}\\
\textsuperscript{\rm 1}University of Science and Technology of China \\
\textsuperscript{\rm 2}Microsoft Research Asia \\
\textsuperscript{\rm 3}Shanghai Jiao Tong University \\
\tt\small hanxinzhu@mail.ustc.edu.cn, tianyuhe@microsoft.com,
\tt\small memory97@sjtu.edu.cn, \\
\tt\small junliangguo@microsoft.com, 
\tt\small chenzhibo@ustc.edu.cn, \tt\small  jiang.bian@microsoft.com
}

\begin{document}

\maketitle

\begin{abstract}
Significant progress has been made in text-to-video generation through the use of powerful generative models and large-scale internet data. However, substantial challenges remain in precisely controlling individual concepts within the generated video, such as the motion and appearance of specific characters and the movement of viewpoints. In this work, we propose a novel paradigm that generates each concept in 3D representation separately and then composes them with priors from Large Language Models (LLM) and 2D diffusion models. Specifically, given an input textual prompt, our scheme consists of three stages: 1) We leverage LLM as the director to first decompose the complex query into several sub-prompts that indicate individual concepts within the video~(\textit{e.g.}, scene, objects, motions), then we let LLM to invoke pre-trained expert models to obtain corresponding 3D representations of concepts. 2) To compose these representations, we prompt multi-modal LLM to produce coarse guidance on the scales and coordinates of trajectories for the objects. 3) To make the generated frames adhere to natural image distribution, we further leverage 2D diffusion priors and use Score Distillation Sampling to refine the composition. Extensive experiments demonstrate that our method can generate high-fidelity videos from text with diverse motion and flexible control over each concept. Project page: \url{https://aka.ms/c3v}.
\end{abstract}

\section{Introduction}
\label{intro}
Benefitting from large-scale data and the advancement of the generative models~\cite{radford2019language,ho2020denoising}, we have witnessed plenty of astonishing results across a wide array of tasks. For example, Large Language Models (LLM) pre-trained on web-scale datasets are revolutionizing machine learning with strong capability of zero-shot learning~\cite{brown2020language} and planning~\cite{schick2024toolformer,huang2023voxposer}, while diffusion models~\cite{dhariwal2021diffusion} empower text-to-image generation with a rapid surge in both quality and diversity~\cite{rombach2022high,ramesh2022hierarchical,saharia2022photorealistic}.

To harness the power of text-to-image models in text-to-video generation, modern solutions directly view video as multiple images. In this way, tremendous efforts have been dedicated to extending text-to-image models with temporal interaction to ensure consistency between frames~\cite{vdm,imagenvideo,blattmann2023align,show1,videocrafter1,girdhar2023emu,gupta2023photorealistic,kondratyuk2023videopoet}. However, generating visual content conditioned on the textual prompt alone struggles to express multiple concepts with precise spatial layout control~\cite{controlnet,feng2024layoutgpt,liu2022compositional}. To tackle this issue, LVD~\cite{lian2023llm} and VideoDirectorGPT~\cite{lin2023videodirectorgpt} propose to first generate spatiotemporal bounding boxes of each object based on the textual prompt with LLM, and then condition the video generation on the obtained layouts. Although rough layout control can be realized, they still have inherent limitations for detailed concept control, \textit{e.g.}, the motion and appearance of specific characters, and the movement of viewpoints.

In nature, our understanding of the world is compositional~\cite{chomsky2014aspects,lake2015human,liu2022compositional}, and the interaction with the world takes place in a 3D. Motivated by this, in contrast to the prior endeavors that \textit{implicitly} learn different concepts in 2D space, we are interested in exploring an alternative solution that \textit{explicitly} composes concepts in 3D space for video generation. To this end, we in particular identify two key technical challenges: 1) Since a textual prompt contains multiple concepts, how to coordinate the generation of various concepts? 2) Given the generated concepts, how to compose them to follow common sense in the real world?

In this work, we introduce text-guided compositional 3D-aware video generation (C3V), a novel paradigm that regards LLM as director and 3D as structural representation for video generation. C3V consists of three main stages: \textbf{1)} Given a textual prompt, to coordinate the generation of various concepts, we leverage LLM to disassemble the input prompt into sub-prompts, where each sub-prompt describes an individual concept, \textit{e.g.}, the scene, objects, and motion. For each concept, a pre-trained expert model is assigned by LLM to generate its corresponding 3D representation (\textit{e.g.}, 3D Gaussians~\cite{kerbl20233d}, SMPL parameters~\cite{loper2015smpl}) according to the textual description. \textbf{2)} To provide coarse instruction for composition (\textit{i.e.}, the scale and trajectory of each object in the scene), we further resort to the priors in multi-modal LLM by querying it with the rendered scene image and the textual goals. However, directly instructing multi-modal LLM to return the scale and trajectory of each object leads to unexpected results, as it is challenging for LLM to estimate visual dynamics. Therefore, we follow a step-by-step reasoning philosophy~\cite{lightman2023let} by representing the object with the bounding boxes and dividing the trajectory estimation into sub-tasks, \textit{i.e.}, estimating the starting points, ending points, and trajectories step-by-step. \textbf{3)} After obtaining the coarse trajectories from the language space, we also propose to refine the scales, rotations, and exact locations with priors from large-scale visual data. Specifically, taking inspiration from DreamFusion~\cite{poole2022dreamfusion}, which proposes to distill generative priors from pre-trained image diffusion models into 3D objects, we employ Score Distillation Sampling (SDS)~\cite{poole2022dreamfusion} to optimize the transformation matrix of each object in 3D space.

Our system has three main advantages: 1) Because each concept is represented by individual 3D representations, it naturally supports flexible control and interaction of each concept. 2) It inherently excels at synthesizing complex and long videos such as drama, etc. 3) The viewpoint is controllable.

Extensive experiments demonstrate that our proposed method can generate 3D-aware videos with diverse motion and high visual quality, even from complex queries that contain multiple concepts and relationships. We also illustrate the flexibility of C3V by editing various concepts of the generated videos. The generated videos are presented on our \href{https://aka.ms/c3v}{project page}. To the best of our knowledge, we make the first attempt towards text-guided compositional 3D-aware video generation. We hope it can inspire further explorations on the interplay between video and 3D generation.

\section{Related Works}

\subsection{Video Generation with LLM}

Recently, there have been substantial efforts in training text-to-video models on large-scale datasets with autoregressive Transformer~\cite{villegas2022phenaki,hong2022cogvideo,kondratyuk2023videopoet} or diffusion models~\cite{vdm,imagenvideo,blattmann2023align,show1,gupta2023photorealistic}. A prominent approach for text-to-video generation is to extend a pre-trained text-to-image model by inserting temporal layers into its architecture, and fine-tuning models on video data. However, although effective, it remains challenging to generate objects with specific attributes or positions. To address this challenge, a series of studies proposed to exploit knowledge from LLM~\cite{chatgpt,gpt4v} to achieve controllable generation~\cite{lian2023llm,feng2024layoutgpt,lin2023videodirectorgpt,su2023motionzero,jain2023peekaboo,zheng2024intelligent}, zero-shot generation~\cite{huang2024free,lu2023flowzero,hong2023direct2v,oh2023mtvg}, or long video generation~\cite{zhuang2024vlogger}. For example, Free-Bloom~\cite{huang2024free} and DirecT2V~\cite{hong2023direct2v} used LLM to transform the input textual prompt into a sequence of sub-prompts that describe each frame. LVD~\cite{lian2023llm} and VideoDirectorGPT~\cite{lin2023videodirectorgpt} employed LLM to generate spatiotemporal bounding boxes to control the object-level dynamics in video generation.

In light of the above success of exploiting LLM to direct video generation in 2D space, we view LLM as a director in 3D, which differs from previous methods not only in terms of technical route but also in benefits: providing free interaction with individual concepts and flexible viewpoint control.

\subsection{Compositional 3D Generation}

% \textcolor{red}{need refine}

Generating 3D assets from textual prompt has garnered significant attention owing to its promising applications in various fields such as AR~\cite{chang2020toward}, VR~\cite{shi2023mvdream}, and autonomous driving~\cite{yue2018lidar}. However, due to the lack of large-scale 3D data, it is challenging to apply 2D generative models to 3D directly. Therefore, building upon Dream Fields~\cite{jain2022zero}, DreamFusion introduced the Score Distillation Sampling (SDS)~\cite{poole2022dreamfusion}, a technique enhancing 3D generation by distilling 2D diffusion priors from pre-trained text-to-image generative models. Motivated by the success of DreamFusion~\cite{poole2022dreamfusion}, dedicated efforts have been made to improve SDS~\cite{wang2024prolificdreamer,huang2023dreamtime,liang2023luciddreamer}. Though achieving remarkable results, these methods struggle to generate scenes with multiple distinct elements. To mitigate this issue, several techniques was proposed to guide 3D generation with additional conditions like layout priors, which we refer to as compositional 3D generation~\cite{po2023compositional,gao2023graphdreamer,zhou2024gala3d}. However, these works still focus on static compositional 3D generation and lack visual dynamic modeling.

Recently, two concurrent works Comp4D~\cite{xu2024comp4d} and TC4D~\cite{bahmani2024tc4d} also achieved compositional 4D generation (\textit{i.e.}, dynamic 3D generation). However, they only considered composition between objects, and the trajectory of these methods is either formulated by kinematics-based equations~\cite{xu2024comp4d} or pre-defined by users~\cite{bahmani2024tc4d}. Differently, we explore 3D-aware video generation with integrated 3D scenes and compose various concepts with priors from both LLM and 2D diffusion models.

\section{Preliminaries}

\subsection{3D Gaussian Splatting}

% \textcolor{red}{to simplify}

3D Gaussian Splatting (3DGS)~\cite{kerbl20233d} has been attracting a lot of interest for novel view synthesis, due to its photorealistic visual quality and real-time rendering. 3DGS utilizes a set of anisotropic ellipsoids (\textit{i.e.}, 3D Gaussians) to encode 3D properties, in which each Gaussian is parameterized by position $\mathbf{\mu} \in \mathbb{R}^3$, covariance $\mathbf{\Sigma} \in \mathbb{R}^{3 \times 3}$ (obtained from scale $\mathbf{s} \in \mathbb{R}^3$ and rotation $\mathbf{r} \in \mathbb{R}^3$), opacity $\alpha \in \mathbb{R}$, and color $\mathbf{c} \in \mathbb{R}^3$.

To render a novel view, 3DGS adopts a tile-based rasterization, where 3D Gaussians are projected onto the image plane as 2D Gaussians. The final color $\mathbf{c}(\mathbf{p})$ of pixel $\mathbf{p}$ is denoted as: 
\begin{equation}
\begin{split}
    \mathbf{c(p)} = \sum \hat{\mathbf{c}}\hat{\sigma}\prod(1-\hat{\sigma}), 
\end{split}
\end{equation}
where $\hat{\textbf{c}}$ and $\hat{\sigma}$ represent the individual color and opacity values of a series of 2D Gaussians contributing to this pixel.  3DGS are then optimized using L1 loss and SSIM~\cite{wang2004image} loss in a per-view optimization manner. Thanks to the nature of modeling 3D scenes explicitly, optimized 3D Gaussians can be easily controlled and edited.

\subsection{Score Distillation Sampling}
Different from text-to-image generation which benefits from a large number of text-image pairs available, text-to-3D generation suffers from a severe lack of data. To mitigate this issue, Score Distillation Sampling (SDS)~\cite{poole2022dreamfusion} was proposed to distill generative priors from pretrained diffusion-based text-to-image models $\phi$. Specifically, for a 3D representation parameterized by $\theta$, SDS is served as a way
to measure the similarity between the rendered images $x = g(\theta)$ and the given textual prompts $y$, where $g$ represents the rendering operation. As a result, the gradients used to update $\theta$ are computed as follows:
\begin{equation}\label{sds}
    \nabla_{\theta}\mathcal{L}_{SDS}(\phi, x = g(\theta)) = \mathbb{E}_{t,\epsilon}[w(t)(\hat{\epsilon}_{\phi}({x}_t;y,t) - \epsilon )],
\end{equation}
where $t$ is the noise level, $\epsilon$ is the ground-truth noise, $w(t)$ is a weighting function, $\hat{\epsilon}_{\phi}$ is the estimated noise given noised images $x_t$ with text embeddings $y$. Please refer to DreamFusion~\cite{poole2022dreamfusion} for details.

\section{Method}

\begin{figure}[htbp]
    \centering
    \includegraphics[width=1\linewidth]
    {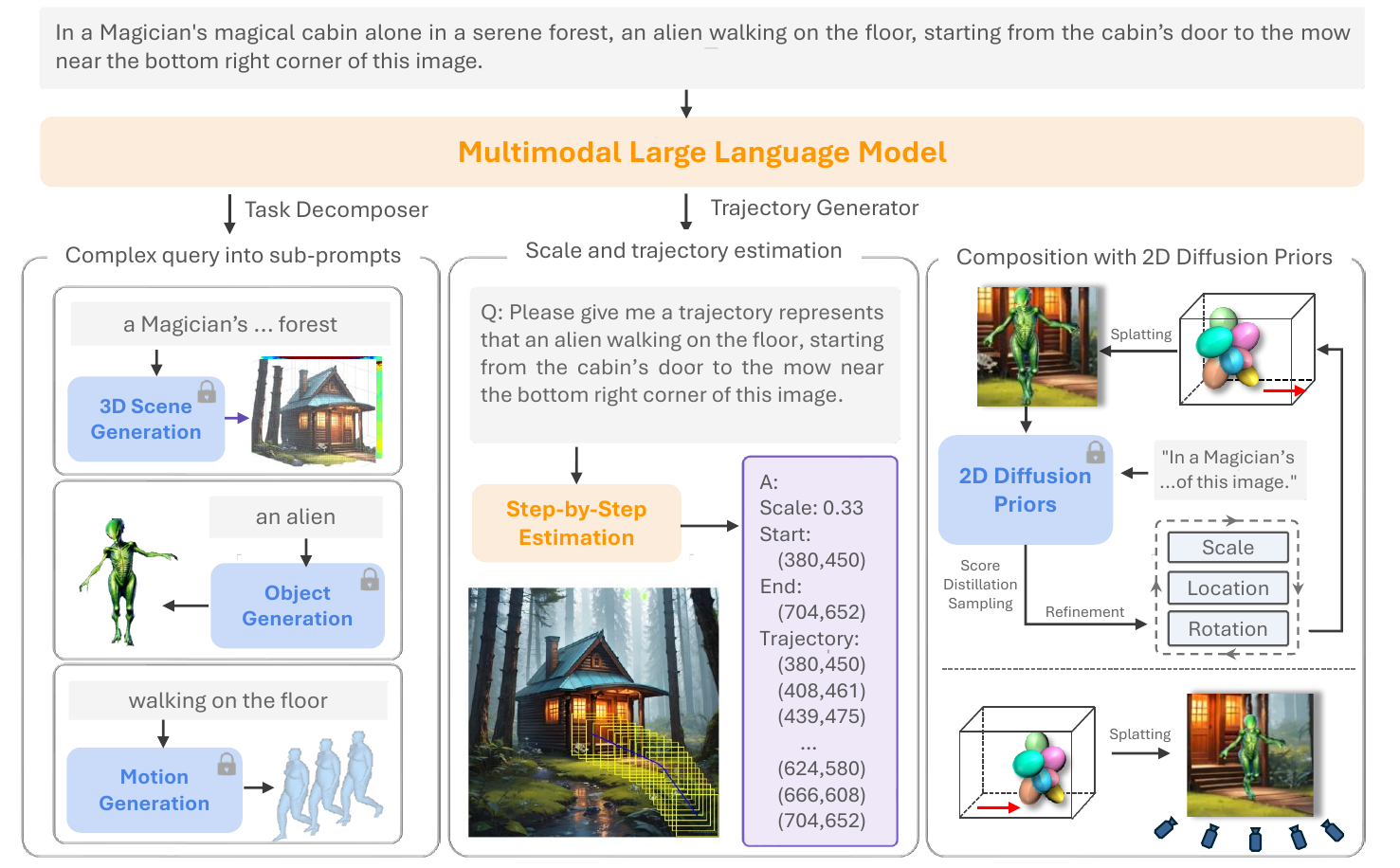}
    \caption{Illustration of our method. It consists of three stages: 1) The input textual prompt is decomposed into individual concepts by the LLM. Then we generate each concept in the form of 3D with the corresponding pre-trained expert model (\textit{left} \& Sec.~\ref{LLM-based Task Decomposition}). 2) We leverage knowledge in multi-modal LLM to estimate the 2D trajectory of objects step-by-step (\textit{middle} \& Sec.~\ref{GPT-4V-based Trajectory Estimation}). 3) After lifting the estimated 2D trajectory into 3D as initialization, we refine the scales, locations, and rotations of objects within the 3D scene using 2D diffusion priors (\textit{right} \& Sec.~\ref{SDS-based Refinement}).}
    \vspace{-3mm}
    \label{fig:framework}
\end{figure}

\paragraph{Overview.}

To achieve text-guided compositional 3D-aware video generation (C3V), we regard LLM as director and 3D as structural representation. To this end, our method consists of three stages. To begin with, we utilize LLMs to decompose the input textual prompts into three sub-prompts, each of which provides a description for generating a corresponding concept (\textit{i.e.}, scene, object, motion, etc.) respectively (Sec.~\ref{LLM-based Task Decomposition}). Subsequently, we leverage multi-modal LLM to obtain coarse-grained scales and trajectories for each animatable object (Sec.~\ref{GPT-4V-based Trajectory Estimation}). Finally, we employ 2D diffusion priors to refine the objects' location, scale, and rotation for a fine-grained composition (Sec.~\ref{SDS-based Refinement}).

\subsection{Task Decomposition with LLM}
\label{LLM-based Task Decomposition}

\paragraph{Task Instructions.}
Given a textual prompt, we invoke LLM (\textit{e.g.}, GPT-4V~\cite{gpt4v}) to decompose it into several sub-prompts. Each sub-prompt describes an individual concept such as the scene, object, and motion. Specifically, for an input prompt $y$, we query LLM with the instruction like: "\textit{Please decompose this prompt into several sub-prompts, each describing the scene, objects in the scene, and the objects' motion.}", from which we obtain the corresponding sub-prompts.
% , \textit{e.g.}, $y_1,y_2$ and $y_3$.
% These sub-prompts are then fed into pre-trained expert models to generate corresponding 3D representations. 

\paragraph{3D Representation.}
After obtaining the sub-prompt for each concept, we aim to generate its corresponding 3D representations using the pre-trained expert models. In this work, we build structural representation on 3DGS~\cite{kerbl20233d}, which is an explicit form and therefore flexible enough to compose or animate. Concerning concepts like motion, our framework can generalize to arbitrary animatable 3D Gaussian-based objects. For simplicity, we take human motion as an instantiation because it is general for various scenarios. In order to obtain diverse human motions, we take a retrieval-augmented approach~\cite{khandelwal2019generalization} to acquire motion in the form of SMPL-X parameters~\cite{pavlakos2019expressive} from large motion libraries~\cite{lin2024motion} according to the motion-related sub-prompt.

\paragraph{Instantiation.}
To illustrate the scheme formally, consider the following example. We have sub-prompts $y_1,y_2$ and $y_3$ that describe scene, object, and motion respectively. Additionally, we have corresponding pre-trained text-guided expert models $\phi_1$, $\phi_2$, and $\phi_3$ that are selected by the LLM. The concept generation can be formulated as follows:
\begin{equation}
\begin{split}
    G_1 = \phi_1(y_1),\ G_2 = \phi_2(y_2, M),\ M = \phi_3(y_3), 
\end{split}
\end{equation}
where $G_1$ and $G_2$ represent the 3D Gaussians, and $M$ means the motions used to drive $G_2$. In the following sections, we will provide details on the composition of the generated concepts.

\subsection{Coarse-grained Trajectory Generation with LLM}
\label{GPT-4V-based Trajectory Estimation}

\begin{figure}[t]
    \centering
    \includegraphics[width=1\linewidth]
    {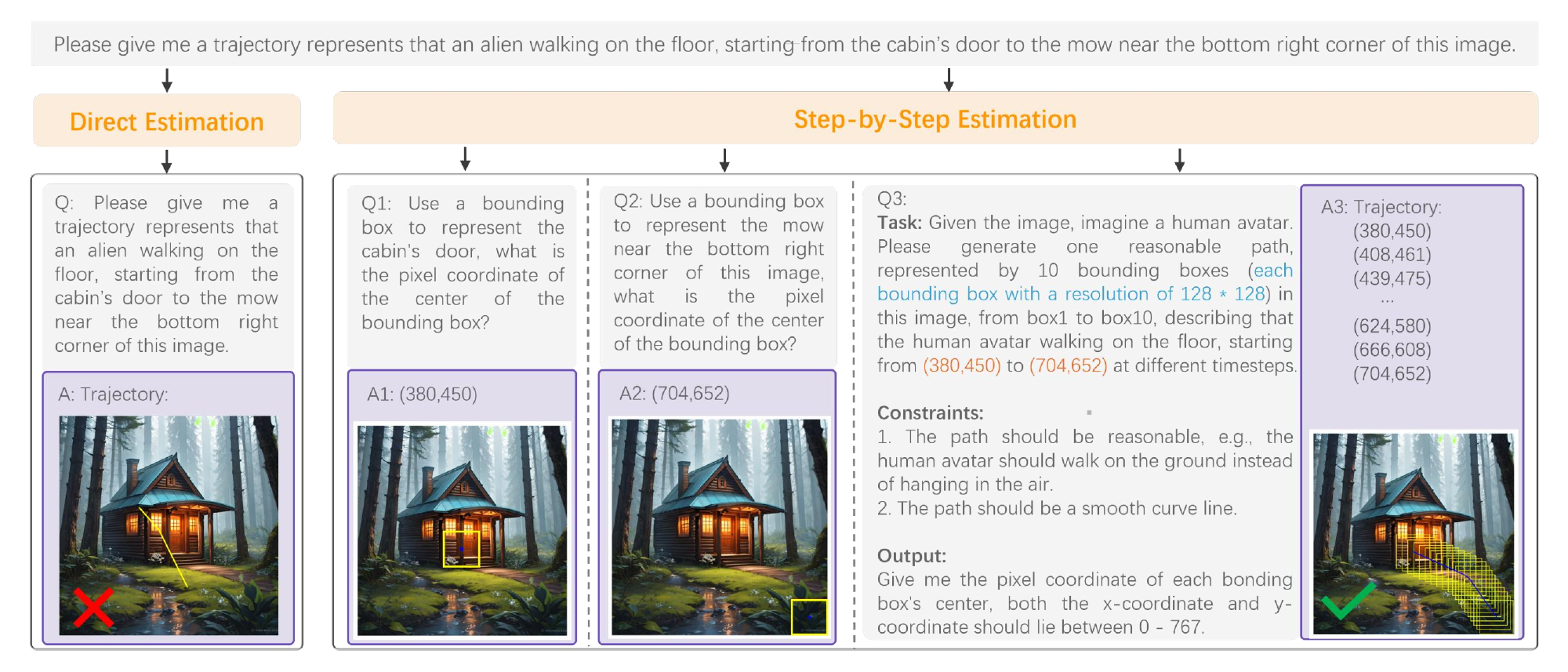}
    \vspace{-5mm}
    \caption{Illustration of coarse-grained trajectory generation with LLM. Instead of querying multi-modal LLM to estimate dynamic trajectory directly, we generate trajectory in a step-by-step manner: estimating the locations of starting and ending points first, then reasoning the path between them.}
    % \vspace{-3mm}
    \label{fig:gpt-4(v) trajectory estimation}
\end{figure}

Given the generated concepts, we aim to compose them into a dynamic 3D representation to render videos that align with the input textual prompt. Achieving this requires scales and trajectories of the objects to indicate their sizes and locations within the scene. To this end, we propose to leverage knowledge encoded in multi-modal LLMs (\textit{i.e}., GPT-4V~\cite{gpt4v}) to provide priors.

For the scale of the object, we find that directly querying GPT-4V with the input prompt and rendered scene image can yield a reasonable estimation of its resolution ($H_{2D}$ and $W_{2D}$). However, this is not the case for trajectory estimation. As demonstrated in Fig.~\ref{fig:gpt-4(v) trajectory estimation}, directly querying GPT-4V for trajectory will lead to a result that deviates conspicuously from common sense. Based on this observation, we conclude two issues: 1) it is too difficult for GPT-4V to generate the trajectory within a single query, as it lacks priors on visual dynamics; 2) since GPT-4V is trained to generate text, it has limitations on imagining visual content.

To mitigate this, we introduce two simple yet effective techniques. 1) Although GPT-4V lacks visual knowledge of the object, we can alleviate this by representing the object as a bounding box with the estimated resolution. 2) We follow a step-by-step reasoning philosophy~\cite{lightman2023let} and propose to let GPT-4V estimate the locations of starting and ending points first, then reason the path between them.

Overall, we can formulate the above process as follows:
% \textcolor{red}{formulation need refine}
\begin{equation}
\begin{split}
&\{L_p^i\}_{i=1}^N = \Phi(y_p, I, S, L_s, L_e),\\
S = &\Phi(y, I), L_s = \Phi(y_s, I), L_e = \Phi(y_e, I),
\end{split} \label{Eq.5}
\end{equation}
where $\Phi$ represents the multi-modal LLM (\textit{i.e.}, GPT-4V), $I$ denotes the rendered scene image, $S$ represents the estimated scale of the object given textual prompt $y$ and $I$, $L_s$ and $L_e$ represent the locations of starting and ending points respectively, $\{L_p^i\}_{i=1}^N$ represent $N$ locations indicating the path between $L_s$ and $L_e$. Notably, all locations (\textit{i.e.}, $L_s$, $L_e$,$\{L_p^i\}_{i=1}^N$ ) are represented by 2D pixel coordinates on $I$.

\subsection{Fine-grained Composition with 2D Diffusion Priors}
\label{SDS-based Refinement}

\paragraph{Lift Trajectory from 2D to 3D.}
In Sec.~\ref{GPT-4V-based Trajectory Estimation}, we obtain the 2D pixel coordinates $L_p^i = (p_x^i,p_y^i)$ of the estimated trajectory. However, 2D trajectory is not enough for composition in 3D space. Therefore, we convert it into corresponding 3D world coordinate $L_{3D}^i = (x^i,y^i,z^i)$. Specifically, we first predict the depth map $D$ of the rendered scene image with a monocular depth estimator~\cite{ranftl2021vision}. Then, we use the depth value of the center point of the lower boundary of the bounding box as the trajectory's depth. As a result, we can transform 2D trajectory into 3D:
\begin{equation}
(x^i,y^i,z^i,1)^T = R^{-1}K^{-1}[(p_x^i + \frac{H_{2D}}{2}, p_y^i, 1)^T\cdot D(p_x^i + \frac{H_{2D}}{2}, p_y^i)] - (\frac{H_{3D}}{2},0,0,0)^T,
\end{equation}
where $R$ and $K$ represent camera extrinsic and intrinsic respectively, $H_{2D}$ and $W_{2D}$ represent the resolution of the 2D bounding box. $H_{3D}$ represent the actual height of the 3D bounding box of this object within the scene.

\paragraph{Composition Refinement with 2D Diffusion Priors.} With the lifted 3D trajectory, we then integrate the object into the scene. However, the trajectory estimated by LLM is still rough and may not obey natural image distribution. To address this, we propose to further refine the object's scale, location, and rotation by distilling generative priors from pre-trained image diffusion models~\cite{rombach2022high} into 3D space. Specifically, we treat the parameters for these attributes as optimizable variables and use SDS (Eq.~\ref{sds}) to improve the fidelity of rendered images. As a result, scale refinement can be formulated as follows:
\begin{equation}\label{Eq.7}
    \nabla_{\hat{S}}\mathcal{L}_{SDS}^{Scale} = \mathbb{E}_{t,\epsilon}[w(t)(\hat{\epsilon}_{\phi}({x}_t(L_{3D}^1, (S + \sigma(\hat{S})\cdot \tau_s - \frac{\tau_s}{2})\cdot G_2);y,t) - \epsilon )],
\end{equation}
where $\hat{S}$ represents the optimizable variable, $S$ represents the scale estimated by GPT-4(V), $\sigma$ means the Sigmoid function, $\tau_s$ is a threshold, $G_2$ represents the 3D gaussians of the object, and $x_t$ is the noised 2D image given $L_{3D}^i$ and scaled $G_2$.

After obtaining a more precise scale, we then refine the locations of the estimated 3D trajectory similarly, where the location refinement is denoted as:
\begin{equation}\label{Eq.8}
    \nabla_{\hat{L^i}}\mathcal{L}_{SDS}^{Location} = \mathbb{E}_{t,\epsilon}[w(t)(\hat{\epsilon}_{\phi}({x}_t(L_{3D}^i + \sigma(\hat{L^i})\cdot \tau_L - \frac{\tau_L}{2}, (S + \sigma(\hat{S})\cdot \tau_s - \frac{\tau_s}{2})\cdot G_2);y,t) - \epsilon )],
\end{equation}
where $\hat{L^i}$ represents the optimizable variable, 
$\tau_L$ is a threshold.

For the rotation of the object at different timesteps, we can directly compute the corresponding rotation matrix, based on the assumption that the object at the current time step should face the location of the object at the next time step. As a result, the rotation matrix $\hat{R^i}$ at time step $i$ can be computed using the following equation:
\begin{equation}
\begin{split}
        &\hat{R^i} = \begin{bmatrix}
t x^2 + c & t xy - zs & t xz + ys \\
t xy + zs & t y^2 + c & t yz - xs \\
t xz - ys & t yz + xs & t z^2 + c
\end{bmatrix}, \\
&t = 1 - c, c = \cos(\theta), s = \sin(\theta),\mathbf{u} = (x,y,z)^T
\end{split}
\end{equation}
where $\theta$ and $\mathbf{u}$ represent the rotation angle and axis obtained through the cross product of $(L_{3D}^{i+1} + \sigma(\hat{L^{i+1}})\cdot \tau_L  - L_{3D}^i - \sigma(\hat{L^i})\cdot \tau_L)$ and $(0,0,1)^T$.

\paragraph{Inference.}

After obtaining individual concepts in the form of 3D and the optimized parameters that indicate how to compose various concepts, we can render the 3D representation into 2D video with flexible camera control in real time~\cite{kerbl20233d}.

\section{Experiments}
% \paragraph{Overview.} In Se.~\ref{main_results}, we provide 

In this section, we instantiate C3V with three concepts: scene, humanoid object, and human motion, to generate 3D-aware video from text. We compare our proposed method with state-of-the-art text-to-4D models (4D-FY~\cite{bahmani20234d}), compositional 4D generation models (Comp4D~\cite{xu2024comp4d}) and text-to-video models (VideoCrafter2~\cite{chen2024videocrafter2}). Videos are available on our \href{https://c3vv.github.io/C3V/}{anonymous project page}.

\paragraph{Implementation Details.}
We use LucidDreamer~\cite{chung2023luciddreamer}, HumanGaussian~\cite{liu2023humangaussian} and Motion-X~\cite{lin2024motion} to generate 3D scenes, humanoid objects and motions respectively. To realize SDS, we utilize Stable Diffusion~\cite{rombach2022high} as the image diffusion model. All the videos of our proposed method are rendered at a resolution of $512 \times 512$ in real time. Please refer to the appendix for more details.
% , \textcolor{red}{with camera views changing}. 

\paragraph{Metrics.}
Following Comp4D~\cite{xu2024comp4d}, we choose Q-Align~\cite{wu2023q} as the referee to measure the quality and aesthetics of the video. The Q-Align score is a number ranging from 1 (worst) to 5 (best) where a higher score indicates a better performance. We also report the CLIP score~\cite{radford2021learning} to measure the alignment between the generated videos and the input texts.

\subsection{Comparison with Competitors}
\label{main_results}

\begin{figure}[t]
    \centering
    \begin{subfigure}[b]{1.0\linewidth}
         \centering 
    \includegraphics[width=1\linewidth]
    {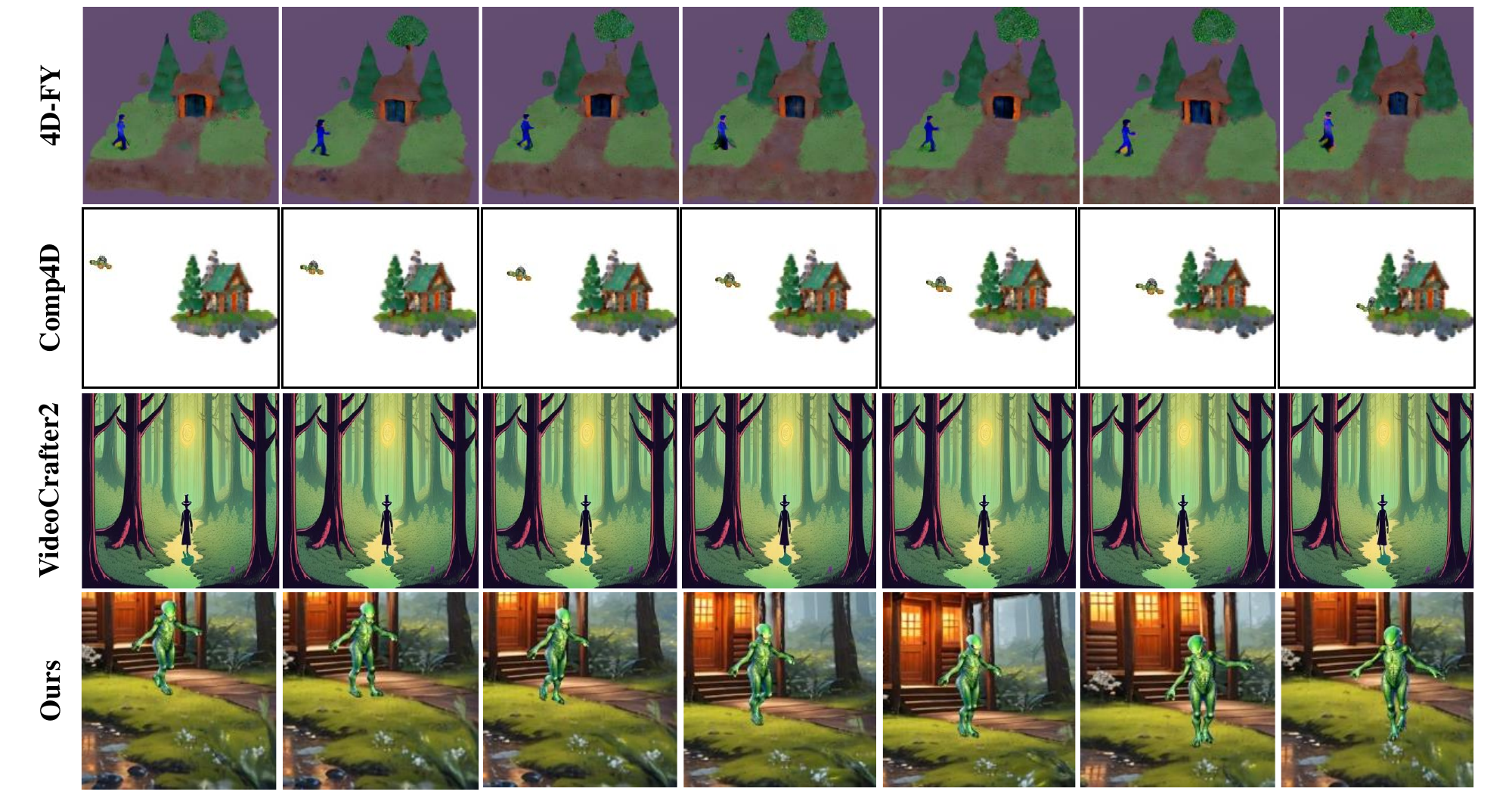}
    \vspace{-5mm}
    \caption{Text prompt: "\textit{In a Magician's magical cabin alone in a serene forest, an alien walking on the floor, starting from the cabin’s door to the mow near the bottom right corner of this image"}.}
    \label{fig:results_1}
    \end{subfigure}
    \begin{subfigure}[b]{1.0\linewidth}
         \centering 
    \includegraphics[width=\linewidth]
    {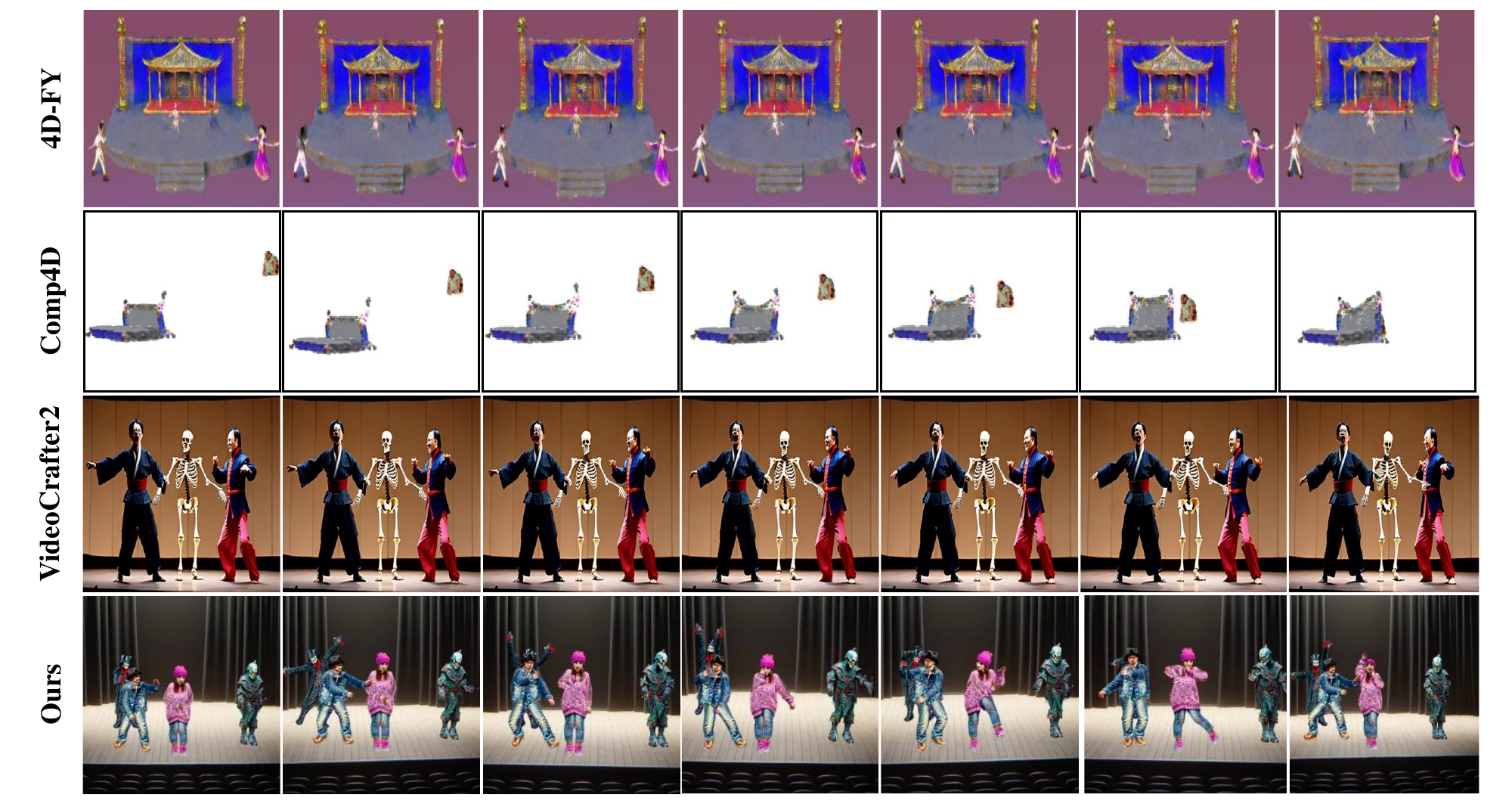}
    \vspace{-5mm}
    \caption{Text prompt: "\textit{Four characters stood on the stage. In front of the stage, a man and a woman are performing Kung Fu and dancing respectively. On the right side of the stage, a skeleton man is dancing, and behind them, a clown is performing"}.}
    \label{fig:results_2}
    \end{subfigure}
    \caption{Qualitative comparisons with baselines. When prompting complex queries, the baseline methods fail to follow the queries in terms of the number of objects and the corresponding motion. In contrast, our method excels in yielding both diverse motion and high visual quality.}
    \vspace{-4mm}
    \label{fig:main results}
\end{figure}

\begin{table}[t]
    \small
    \centering
    % \captionsetup{justification=centering}
    \caption{Quantative comparisons with competitors. Our method consistently outperforms all baseline methods in terms of both the video quality and the alignment with textual prompts.}
    \label{tab:main_results}
    \begin{tabular}{l|cccc}
        \toprule
        {Metric} & {4D-FY~\cite{bahmani20234d}} & {Comp4D~\cite{xu2024comp4d}} & {VideoCrafter2~\cite{chen2024videocrafter2}} & Ours \\
        \midrule
        QAlign-img-quality $\uparrow$~\cite{wu2023q} & 1.681 & 1.687 & 3.839 & \textbf{4.030} \\
        QAlign-img-aesthetic$\uparrow$~\cite{wu2023q} & 1.475 & 1.258 & 3.199 & \textbf{3.471} \\
        QAlign-vid-quality$\uparrow$~\cite{wu2023q} & 2.154 & 2.142 & 3.868 & \textbf{4.112} \\
        QAlign-vid-aesthetic$\uparrow$~\cite{wu2023q} & 1.580 & 1.425 & 3.159 & \textbf{3.723} \\
        \midrule
        CLIP Score$\uparrow$~\cite{radford2021learning} & 30.47 & 27.50 & 35.20 & \textbf{38.36} \\
        \bottomrule
    \end{tabular}
\end{table}

In Fig.~\ref{fig:main results}, we conduct a comparative analysis of our method against 4D-FY~\cite{bahmani20234d}, Comp4D~\cite{xu2024comp4d}, and VideoCrafter2~\cite{chen2024videocrafter2} with the same textual prompt. It can be observed that all three baselines fail to provide diverse motion from the textual prompt, while our method excels in yielding large motion and high visual quality. For example, our scheme successfully obeys the complex query in terms of the number of objects and the corresponding motion. In addition, since 4D-FY and Comp4D focus on object-centric generation, they fail to generate videos with natural backgrounds. In Tab.~\ref{tab:main_results}, we perform quantitative comparisons by utilizing Q-Align Score~\cite{wu2023q} and CLIP Score~\cite{radford2021learning} to assess the quality of generated videos. Our method consistently outperforms the baseline models in terms of both the video quality and the alignment with textual prompts. More results are available in the appendix.

\subsection{Ablation Studies}

\paragraph{Ablations on Trajectory Estimation with Multi-modal LLM.}

\begin{figure}[t]
    \centering
    \includegraphics[width=1\linewidth]
    {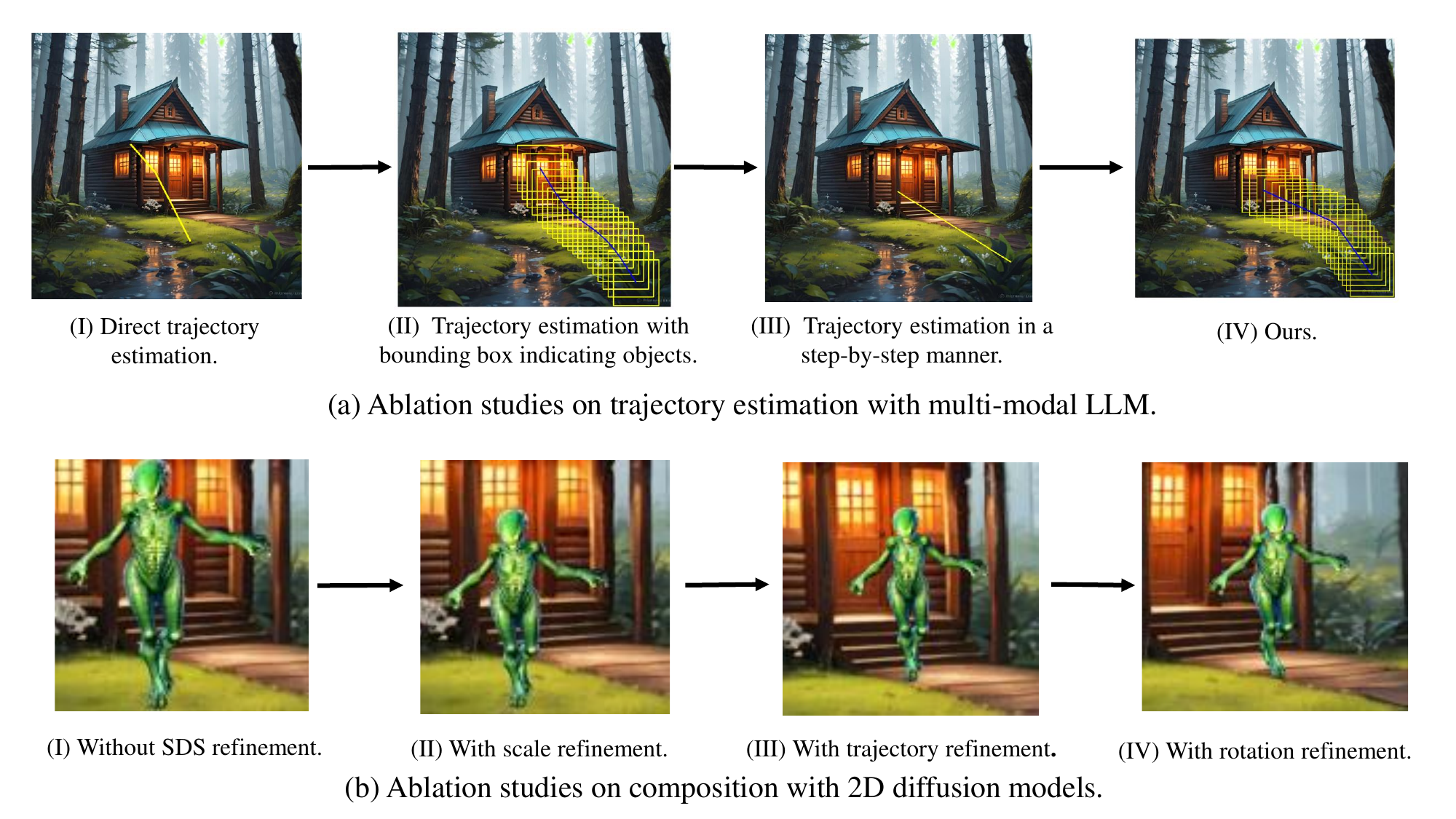}
    \vspace{-6mm}
    \caption{Ablation studies on framework design. Each ablation is prompted with the same text.}
    \label{fig:ablation_result}
\end{figure}

As shown in Fig.~\ref{fig:ablation_result}(a)(I), a direct prompt of GPT-4V will lead to obvious unsatisfactory trajectory estimation. When only depending on bounding boxes to indicate the location of objects within the scene (Fig.~\ref{fig:ablation_result}(a)(II)), though a roughly better trajectory can be achieved, it still leads to unreasonable results, such as several floating bounding boxes. Similarly, using only the step-by-step estimation strategy described in Sec.~\ref{GPT-4V-based Trajectory Estimation} typically results in a trajectory that is merely a simple straight line connecting the starting and ending points (Fig.~\ref{fig:ablation_result}(a)(III)). With both of the two techniques, we can achieve the best performance, with a more reasonable and smooth trajectory (Fig.~\ref{fig:ablation_result}(a)(IV)).

\paragraph{Ablations on Composition with 2D Diffusion Models.}

To figure out whether it is necessary to conduct fine-grained composition with 2D generative priors, we gradually refine the scales, locations, and rotations with SDS and visualize the results in Fig.~\ref{fig:ablation_result}(b). All results are generated with the same textual prompt: "\textit{An alien walking on the floor in front of the cabin's door.}". It shows that when we optimize the attributes with SDS, we can obtain consistently improved performance with a reasonable scale (Fig.~\ref{fig:ablation_result}(b)(II), accurate locations that are aligned with the input prompt (Fig.~\ref{fig:ablation_result}(b)(III), and orientation that accords with common sense (Fig.~\ref{fig:ablation_result}(b)(IV)).

\subsection{Applications on Controllable Generation}

\begin{figure}[t]
    \centering
    \includegraphics[width=1\linewidth]
    {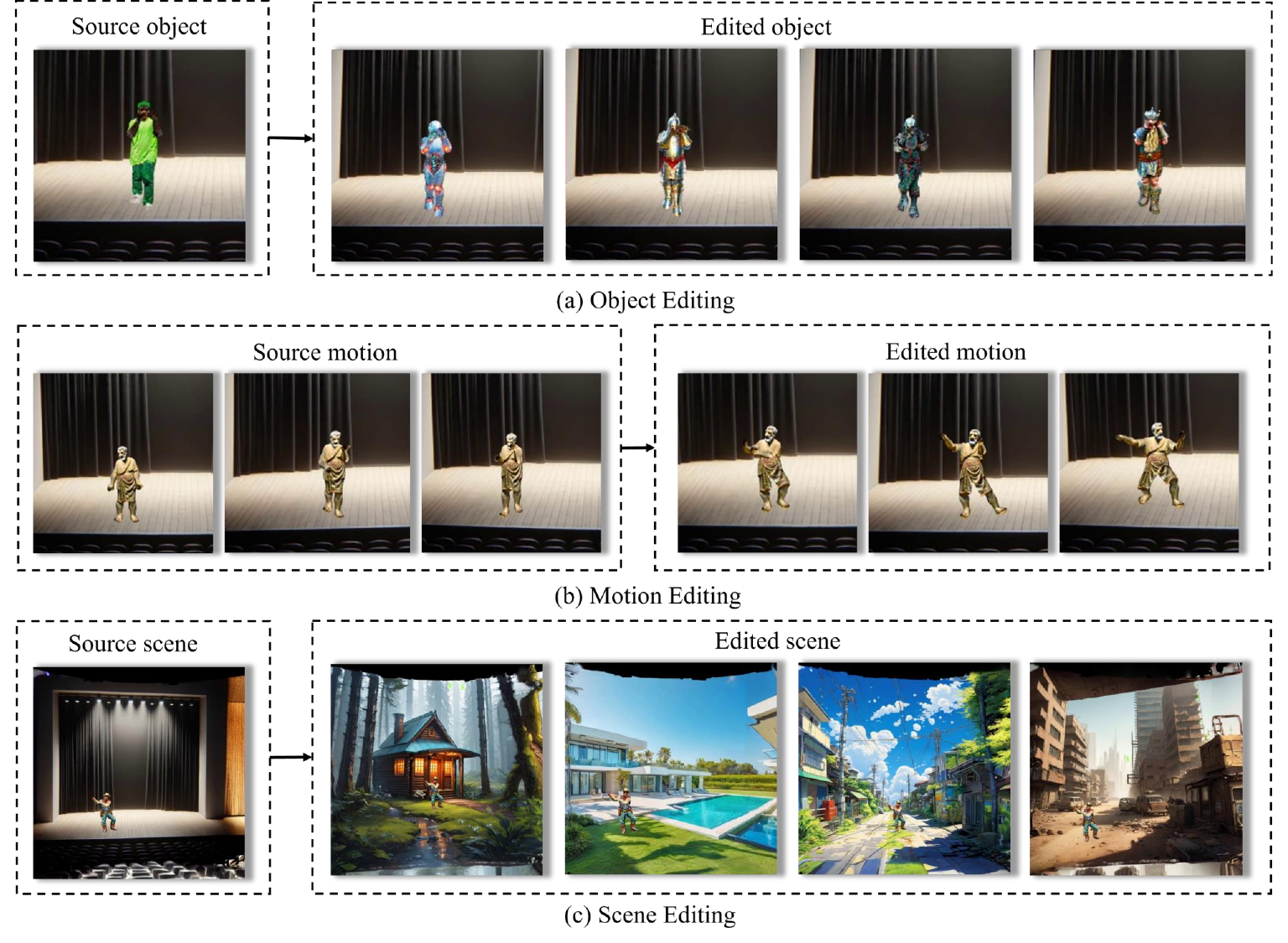}
    \vspace{-5.5mm}
    \caption{Our method offers flexible control of individual concepts. We demonstrate this by editing different concepts: the appearance and motion of the actors, and the scenes.}
    \label{fig:editing_result}
\end{figure}

Due to our underlying 3D structural representation, our scheme has the natural merits of editing individual concepts. We illustrate this character in Fig.~\ref{fig:editing_result} by editing three different concepts: the appearance and motion of the actors, and the scenes. For the appearance and motion of the actor, we can seamlessly replace them in a zero-shot manner according to the textual prompt (Fig.\ref{fig:editing_result}(a)(b)), while this is still challenging for implicit models~\cite{yan2023motion,bai2024uniedit}. For scene editing, to ensure a smooth composition of objects within the target scene, we re-estimate the trajectory of the objects given the target scene. Kindly refer to appendix for more results.

\section{Conclusion}
In this paper, we present a novel paradigm for 3D-aware video generation by conceptualizing videos as compositions of independent concepts represented in 3D space. To this end, we leverage LLM as director to decompose the input textual prompts into individual concepts and then invoke pre-trained expert models to generate them separately. To compose various concepts, we first prompt multi-modal LLM in a step-by-step manner to provide coarse guidance on the scale and trajectory of objects, then refine the composition with 2D generative priors. We verify our scheme in different scenarios, demonstrating its superiority over the baseline methods.

\paragraph{Limitations and Future Works.}
Although we demonstrate promising results in 3D-aware video generation, there still are limitations to be improved in the future. First, our framework is instantiated with limited concepts in this work, \textit{i.e.}, scene, humanoid object, and human motion. It is exciting to generalize the framework to more concepts like animals, vehicles, etc. Second, the composition between concepts is conducted with priors from LLM and 2D diffusion priors in our method. However, it is still interesting to introduce physically grounded dynamics into 3D representation~\cite{xie2023physgaussian}. Third, though our method is naturally suitable for maintaining the consistency of actors across different scenes, it still needs further exploration on long video generation with multiple scenes, \textit{e.g.}, a full-length film.

\paragraph{Ethics Statement.}
C3V is exclusively a research initiative with no current plans for product integration or public access. We are committed to adhering to Microsoft AI principles during the ongoing development of our models. The model is trained on AI-generated content, which has been thoroughly reviewed to ensure that they do not include personally identifiable information or offensive content. Nonetheless, as these generated data are sourced from the Internet, there may still be inherent biases. To address this, we have implemented a rigorous filtering process on the data to minimize the potential for the model to generate inappropriate content.

\bibliographystyle{unsrt}
\bibliography{neurips_2024}

\begin{thebibliography}{10}

\bibitem{radford2019language}
Alec Radford, Jeffrey Wu, Rewon Child, David Luan, Dario Amodei, Ilya Sutskever, et~al.
\newblock Language models are unsupervised multitask learners.
\newblock {\em OpenAI blog}, 1(8):9, 2019.

\bibitem{ho2020denoising}
Jonathan Ho, Ajay Jain, and Pieter Abbeel.
\newblock Denoising diffusion probabilistic models.
\newblock {\em Advances in Neural Information Processing Systems (NeurIPS)}, 33:6840--6851, 2020.

\bibitem{brown2020language}
Tom Brown, Benjamin Mann, Nick Ryder, Melanie Subbiah, Jared~D Kaplan, Prafulla Dhariwal, Arvind Neelakantan, Pranav Shyam, Girish Sastry, Amanda Askell, et~al.
\newblock Language models are few-shot learners.
\newblock {\em Advances in Neural Information Processing Systems (NeurIPS)}, 33:1877--1901, 2020.

\bibitem{schick2024toolformer}
Timo Schick, Jane Dwivedi-Yu, Roberto Dess{\`\i}, Roberta Raileanu, Maria Lomeli, Eric Hambro, Luke Zettlemoyer, Nicola Cancedda, and Thomas Scialom.
\newblock Toolformer: Language models can teach themselves to use tools.
\newblock {\em Advances in Neural Information Processing Systems}, 36, 2024.

\bibitem{huang2023voxposer}
Wenlong Huang, Chen Wang, Ruohan Zhang, Yunzhu Li, Jiajun Wu, and Li~Fei-Fei.
\newblock Voxposer: Composable 3d value maps for robotic manipulation with language models.
\newblock In {\em Conference on Robot Learning}, pages 540--562. PMLR, 2023.

\bibitem{dhariwal2021diffusion}
Prafulla Dhariwal and Alexander Nichol.
\newblock Diffusion models beat gans on image synthesis.
\newblock {\em Advances in Neural Information Processing Systems (NeurIPS)}, 34:8780--8794, 2021.

\bibitem{rombach2022high}
Robin Rombach, Andreas Blattmann, Dominik Lorenz, Patrick Esser, and Bj{\"o}rn Ommer.
\newblock High-resolution image synthesis with latent diffusion models.
\newblock In {\em Proceedings of the IEEE/CVF Conference on Computer Vision and Pattern Recognition (CVPR)}, pages 10684--10695, 2022.

\bibitem{ramesh2022hierarchical}
Aditya Ramesh, Prafulla Dhariwal, Alex Nichol, Casey Chu, and Mark Chen.
\newblock Hierarchical text-conditional image generation with clip latents.
\newblock {\em arXiv preprint arXiv:2204.06125}, 2022.

\bibitem{saharia2022photorealistic}
Chitwan Saharia, William Chan, Saurabh Saxena, Lala Li, Jay Whang, Emily~L Denton, Kamyar Ghasemipour, Raphael Gontijo~Lopes, Burcu Karagol~Ayan, Tim Salimans, et~al.
\newblock Photorealistic text-to-image diffusion models with deep language understanding.
\newblock {\em Advances in Neural Information Processing Systems (NeurIPS)}, 35:36479--36494, 2022.

\bibitem{vdm}
Jonathan Ho, Tim Salimans, Alexey Gritsenko, William Chan, Mohammad Norouzi, and David~J Fleet.
\newblock Video diffusion models.
\newblock {\em arXiv:2204.03458}, 2022.

\bibitem{imagenvideo}
Jonathan Ho, William Chan, Chitwan Saharia, Jay Whang, Ruiqi Gao, Alexey Gritsenko, Diederik~P Kingma, Ben Poole, Mohammad Norouzi, David~J Fleet, et~al.
\newblock Imagen video: High definition video generation with diffusion models.
\newblock {\em arXiv preprint arXiv:2210.02303}, 2022.

\bibitem{blattmann2023align}
Andreas Blattmann, Robin Rombach, Huan Ling, Tim Dockhorn, Seung~Wook Kim, Sanja Fidler, and Karsten Kreis.
\newblock Align your latents: High-resolution video synthesis with latent diffusion models.
\newblock In {\em Proceedings of the IEEE/CVF Conference on Computer Vision and Pattern Recognition (CVPR)}, pages 22563--22575, 2023.

\bibitem{show1}
David~Junhao Zhang, Jay~Zhangjie Wu, Jia-Wei Liu, Rui Zhao, Lingmin Ran, Yuchao Gu, Difei Gao, and Mike~Zheng Shou.
\newblock Show-1: Marrying pixel and latent diffusion models for text-to-video generation.
\newblock {\em arXiv preprint arXiv:2309.15818}, 2023.

\bibitem{videocrafter1}
Haoxin Chen, Menghan Xia, Yingqing He, Yong Zhang, Xiaodong Cun, Shaoshu Yang, Jinbo Xing, Yaofang Liu, Qifeng Chen, Xintao Wang, et~al.
\newblock Videocrafter1: Open diffusion models for high-quality video generation.
\newblock {\em arXiv preprint arXiv:2310.19512}, 2023.

\bibitem{girdhar2023emu}
Rohit Girdhar, Mannat Singh, Andrew Brown, Quentin Duval, Samaneh Azadi, Sai~Saketh Rambhatla, Akbar Shah, Xi~Yin, Devi Parikh, and Ishan Misra.
\newblock Emu video: Factorizing text-to-video generation by explicit image conditioning.
\newblock {\em arXiv preprint arXiv:2311.10709}, 2023.

\bibitem{gupta2023photorealistic}
Agrim Gupta, Lijun Yu, Kihyuk Sohn, Xiuye Gu, Meera Hahn, Li~Fei-Fei, Irfan Essa, Lu~Jiang, and Jos{\'e} Lezama.
\newblock Photorealistic video generation with diffusion models.
\newblock {\em arXiv preprint arXiv:2312.06662}, 2023.

\bibitem{kondratyuk2023videopoet}
Dan Kondratyuk, Lijun Yu, Xiuye Gu, Jos{\'e} Lezama, Jonathan Huang, Rachel Hornung, Hartwig Adam, Hassan Akbari, Yair Alon, Vighnesh Birodkar, et~al.
\newblock Videopoet: A large language model for zero-shot video generation.
\newblock {\em arXiv preprint arXiv:2312.14125}, 2023.

\bibitem{controlnet}
Lvmin Zhang, Anyi Rao, and Maneesh Agrawala.
\newblock Adding conditional control to text-to-image diffusion models.
\newblock In {\em Proceedings of the IEEE/CVF International Conference on Computer Vision}, pages 3836--3847, 2023.

\bibitem{feng2024layoutgpt}
Weixi Feng, Wanrong Zhu, Tsu-jui Fu, Varun Jampani, Arjun Akula, Xuehai He, Sugato Basu, Xin~Eric Wang, and William~Yang Wang.
\newblock Layoutgpt: Compositional visual planning and generation with large language models.
\newblock {\em Advances in Neural Information Processing Systems}, 36, 2024.

\bibitem{liu2022compositional}
Nan Liu, Shuang Li, Yilun Du, Antonio Torralba, and Joshua~B Tenenbaum.
\newblock Compositional visual generation with composable diffusion models.
\newblock In {\em European Conference on Computer Vision}, pages 423--439. Springer, 2022.

\bibitem{lian2023llm}
Long Lian, Baifeng Shi, Adam Yala, Trevor Darrell, and Boyi Li.
\newblock Llm-grounded video diffusion models.
\newblock In {\em International Conference on Learning Representations}, 2024.

\bibitem{lin2023videodirectorgpt}
Han Lin, Abhay Zala, Jaemin Cho, and Mohit Bansal.
\newblock Videodirectorgpt: Consistent multi-scene video generation via llm-guided planning.
\newblock {\em arXiv preprint arXiv:2309.15091}, 2023.

\bibitem{chomsky2014aspects}
Noam Chomsky.
\newblock {\em Aspects of the Theory of Syntax}.
\newblock Number~11. MIT press, 2014.

\bibitem{lake2015human}
Brenden~M Lake, Ruslan Salakhutdinov, and Joshua~B Tenenbaum.
\newblock Human-level concept learning through probabilistic program induction.
\newblock {\em Science}, 350(6266):1332--1338, 2015.

\bibitem{kerbl20233d}
Bernhard Kerbl, Georgios Kopanas, Thomas Leimk{\"u}hler, and George Drettakis.
\newblock 3d gaussian splatting for real-time radiance field rendering.
\newblock {\em ACM Transactions on Graphics}, 42(4):1--14, 2023.

\bibitem{loper2015smpl}
Matthew Loper, Naureen Mahmood, Javier Romero, Gerard Pons-Moll, and Michael~J Black.
\newblock Smpl: a skinned multi-person linear model.
\newblock {\em ACM Transactions on Graphics (TOG)}, 34(6):1--16, 2015.

\bibitem{lightman2023let}
Hunter Lightman, Vineet Kosaraju, Yuri Burda, Harrison Edwards, Bowen Baker, Teddy Lee, Jan Leike, John Schulman, Ilya Sutskever, and Karl Cobbe.
\newblock Let's verify step by step.
\newblock In {\em The Twelfth International Conference on Learning Representations}, 2024.

\bibitem{poole2022dreamfusion}
Ben Poole, Ajay Jain, Jonathan~T Barron, and Ben Mildenhall.
\newblock Dreamfusion: Text-to-3d using 2d diffusion.
\newblock In {\em The Eleventh International Conference on Learning Representations}, 2023.

\bibitem{villegas2022phenaki}
Ruben Villegas, Mohammad Babaeizadeh, Pieter-Jan Kindermans, Hernan Moraldo, Han Zhang, Mohammad~Taghi Saffar, Santiago Castro, Julius Kunze, and Dumitru Erhan.
\newblock Phenaki: Variable length video generation from open domain textual descriptions.
\newblock In {\em International Conference on Learning Representations}, 2023.

\bibitem{hong2022cogvideo}
Wenyi Hong, Ming Ding, Wendi Zheng, Xinghan Liu, and Jie Tang.
\newblock Cogvideo: Large-scale pretraining for text-to-video generation via transformers.
\newblock In {\em International Conference on Learning Representations}, 2023.

\bibitem{chatgpt}
OpenAI.
\newblock Chatgpt.
\newblock {\em https://openai.com/chatgpt}, 2022.

\bibitem{gpt4v}
OpenAI.
\newblock Gpt-4v(ision) system card.
\newblock {\em https://openai.com/index/gpt-4v-system-card}, 2023.

\bibitem{su2023motionzero}
Sitong Su, Litao Guo, Lianli Gao, Hengtao Shen, and Jingkuan Song.
\newblock Motionzero: Exploiting motion priors for zero-shot text-to-video generation.
\newblock {\em arXiv preprint arXiv:2311.16635}, 2023.

\bibitem{jain2023peekaboo}
Yash Jain, Anshul Nasery, Vibhav Vineet, and Harkirat Behl.
\newblock Peekaboo: Interactive video generation via masked-diffusion.
\newblock {\em arXiv preprint arXiv:2312.07509}, 2023.

\bibitem{zheng2024intelligent}
Sixiao Zheng, Jingyang Huo, Yu~Wang, and Yanwei Fu.
\newblock Intelligent director: An automatic framework for dynamic visual composition using chatgpt.
\newblock {\em arXiv preprint arXiv:2402.15746}, 2024.

\bibitem{huang2024free}
Hanzhuo Huang, Yufan Feng, Cheng Shi, Lan Xu, Jingyi Yu, and Sibei Yang.
\newblock Free-bloom: Zero-shot text-to-video generator with llm director and ldm animator.
\newblock {\em Advances in Neural Information Processing Systems}, 36, 2024.

\bibitem{lu2023flowzero}
Yu~Lu, Linchao Zhu, Hehe Fan, and Yi~Yang.
\newblock Flowzero: Zero-shot text-to-video synthesis with llm-driven dynamic scene syntax.
\newblock {\em arXiv preprint arXiv:2311.15813}, 2023.

\bibitem{hong2023direct2v}
Susung Hong, Junyoung Seo, Heeseong Shin, Sunghwan Hong, and Seungryong Kim.
\newblock Direct2v: Large language models are frame-level directors for zero-shot text-to-video generation.
\newblock {\em arXiv preprint arXiv:2305.14330}, 2023.

\bibitem{oh2023mtvg}
Gyeongrok Oh, Jaehwan Jeong, Sieun Kim, Wonmin Byeon, Jinkyu Kim, Sungwoong Kim, Hyeokmin Kwon, and Sangpil Kim.
\newblock Mtvg: Multi-text video generation with text-to-video models.
\newblock {\em arXiv preprint arXiv:2312.04086}, 2023.

\bibitem{zhuang2024vlogger}
Shaobin Zhuang, Kunchang Li, Xinyuan Chen, Yaohui Wang, Ziwei Liu, Yu~Qiao, and Yali Wang.
\newblock Vlogger: Make your dream a vlog.
\newblock {\em arXiv preprint arXiv:2401.09414}, 2024.

\bibitem{chang2020toward}
Chenliang Chang, Kiseung Bang, Gordon Wetzstein, Byoungho Lee, and Liang Gao.
\newblock Toward the next-generation vr/ar optics: a review of holographic near-eye displays from a human-centric perspective.
\newblock {\em Optica}, 7(11):1563--1578, 2020.

\bibitem{shi2023mvdream}
Yichun Shi, Peng Wang, Jianglong Ye, Mai Long, Kejie Li, and Xiao Yang.
\newblock Mvdream: Multi-view diffusion for 3d generation.
\newblock {\em arXiv preprint arXiv:2308.16512}, 2023.

\bibitem{yue2018lidar}
Xiangyu Yue, Bichen Wu, Sanjit~A Seshia, Kurt Keutzer, and Alberto~L Sangiovanni-Vincentelli.
\newblock A lidar point cloud generator: from a virtual world to autonomous driving.
\newblock In {\em Proceedings of the 2018 ACM on International Conference on Multimedia Retrieval}, pages 458--464, 2018.

\bibitem{jain2022zero}
Ajay Jain, Ben Mildenhall, Jonathan~T Barron, Pieter Abbeel, and Ben Poole.
\newblock Zero-shot text-guided object generation with dream fields.
\newblock In {\em Proceedings of the IEEE/CVF conference on computer vision and pattern recognition}, pages 867--876, 2022.

\bibitem{wang2024prolificdreamer}
Zhengyi Wang, Cheng Lu, Yikai Wang, Fan Bao, Chongxuan Li, Hang Su, and Jun Zhu.
\newblock Prolificdreamer: High-fidelity and diverse text-to-3d generation with variational score distillation.
\newblock {\em Advances in Neural Information Processing Systems}, 36, 2024.

\bibitem{huang2023dreamtime}
Yukun Huang, Jianan Wang, Yukai Shi, Xianbiao Qi, Zheng-Jun Zha, and Lei Zhang.
\newblock Dreamtime: An improved optimization strategy for text-to-3d content creation.
\newblock {\em arXiv preprint arXiv:2306.12422}, 2023.

\bibitem{liang2023luciddreamer}
Yixun Liang, Xin Yang, Jiantao Lin, Haodong Li, Xiaogang Xu, and Yingcong Chen.
\newblock Luciddreamer: Towards high-fidelity text-to-3d generation via interval score matching.
\newblock {\em arXiv preprint arXiv:2311.11284}, 2023.

\bibitem{po2023compositional}
Ryan Po and Gordon Wetzstein.
\newblock Compositional 3d scene generation using locally conditioned diffusion.
\newblock {\em arXiv preprint arXiv:2303.12218}, 2023.

\bibitem{gao2023graphdreamer}
Gege Gao, Weiyang Liu, Anpei Chen, Andreas Geiger, and Bernhard Sch{\"o}lkopf.
\newblock Graphdreamer: Compositional 3d scene synthesis from scene graphs.
\newblock {\em arXiv preprint arXiv:2312.00093}, 2023.

\bibitem{zhou2024gala3d}
Xiaoyu Zhou, Xingjian Ran, Yajiao Xiong, Jinlin He, Zhiwei Lin, Yongtao Wang, Deqing Sun, and Ming-Hsuan Yang.
\newblock Gala3d: Towards text-to-3d complex scene generation via layout-guided generative gaussian splatting.
\newblock {\em arXiv preprint arXiv:2402.07207}, 2024.

\bibitem{xu2024comp4d}
Dejia Xu, Hanwen Liang, Neel~P Bhatt, Hezhen Hu, Hanxue Liang, Konstantinos~N Plataniotis, and Zhangyang Wang.
\newblock Comp4d: Llm-guided compositional 4d scene generation.
\newblock {\em arXiv preprint arXiv:2403.16993}, 2024.

\bibitem{bahmani2024tc4d}
Sherwin Bahmani, Xian Liu, Yifan Wang, Ivan Skorokhodov, Victor Rong, Ziwei Liu, Xihui Liu, Jeong~Joon Park, Sergey Tulyakov, Gordon Wetzstein, et~al.
\newblock Tc4d: Trajectory-conditioned text-to-4d generation.
\newblock {\em arXiv preprint arXiv:2403.17920}, 2024.

\bibitem{wang2004image}
Zhou Wang, Alan~C Bovik, Hamid~R Sheikh, and Eero~P Simoncelli.
\newblock Image quality assessment: from error visibility to structural similarity.
\newblock {\em IEEE transactions on image processing}, 13(4):600--612, 2004.

\bibitem{khandelwal2019generalization}
Urvashi Khandelwal, Omer Levy, Dan Jurafsky, Luke Zettlemoyer, and Mike Lewis.
\newblock Generalization through memorization: Nearest neighbor language models.
\newblock In {\em International Conference on Learning Representations}, 2020.

\bibitem{pavlakos2019expressive}
Georgios Pavlakos, Vasileios Choutas, Nima Ghorbani, Timo Bolkart, Ahmed~AA Osman, Dimitrios Tzionas, and Michael~J Black.
\newblock Expressive body capture: 3d hands, face, and body from a single image.
\newblock In {\em Proceedings of the IEEE/CVF conference on computer vision and pattern recognition}, pages 10975--10985, 2019.

\bibitem{lin2024motion}
Jing Lin, Ailing Zeng, Shunlin Lu, Yuanhao Cai, Ruimao Zhang, Haoqian Wang, and Lei Zhang.
\newblock Motion-x: A large-scale 3d expressive whole-body human motion dataset.
\newblock {\em Advances in Neural Information Processing Systems}, 36, 2024.

\bibitem{ranftl2021vision}
Ren{\'e} Ranftl, Alexey Bochkovskiy, and Vladlen Koltun.
\newblock Vision transformers for dense prediction.
\newblock In {\em Proceedings of the IEEE/CVF international conference on computer vision}, pages 12179--12188, 2021.

\bibitem{bahmani20234d}
Sherwin Bahmani, Ivan Skorokhodov, Victor Rong, Gordon Wetzstein, Leonidas Guibas, Peter Wonka, Sergey Tulyakov, Jeong~Joon Park, Andrea Tagliasacchi, and David~B Lindell.
\newblock 4d-fy: Text-to-4d generation using hybrid score distillation sampling.
\newblock {\em arXiv preprint arXiv:2311.17984}, 2023.

\bibitem{chen2024videocrafter2}
Haoxin Chen, Yong Zhang, Xiaodong Cun, Menghan Xia, Xintao Wang, Chao Weng, and Ying Shan.
\newblock Videocrafter2: Overcoming data limitations for high-quality video diffusion models.
\newblock {\em arXiv preprint arXiv:2401.09047}, 2024.

\bibitem{chung2023luciddreamer}
Jaeyoung Chung, Suyoung Lee, Hyeongjin Nam, Jaerin Lee, and Kyoung~Mu Lee.
\newblock Luciddreamer: Domain-free generation of 3d gaussian splatting scenes.
\newblock {\em arXiv preprint arXiv:2311.13384}, 2023.

\bibitem{liu2023humangaussian}
Xian Liu, Xiaohang Zhan, Jiaxiang Tang, Ying Shan, Gang Zeng, Dahua Lin, Xihui Liu, and Ziwei Liu.
\newblock Humangaussian: Text-driven 3d human generation with gaussian splatting.
\newblock {\em arXiv preprint arXiv:2311.17061}, 2023.

\bibitem{wu2023q}
Haoning Wu, Zicheng Zhang, Weixia Zhang, Chaofeng Chen, Liang Liao, Chunyi Li, Yixuan Gao, Annan Wang, Erli Zhang, Wenxiu Sun, et~al.
\newblock Q-align: Teaching lmms for visual scoring via discrete text-defined levels.
\newblock {\em arXiv preprint arXiv:2312.17090}, 2023.

\bibitem{radford2021learning}
Alec Radford, Jong~Wook Kim, Chris Hallacy, Aditya Ramesh, Gabriel Goh, Sandhini Agarwal, Girish Sastry, Amanda Askell, Pamela Mishkin, Jack Clark, et~al.
\newblock Learning transferable visual models from natural language supervision.
\newblock In {\em International conference on machine learning}, pages 8748--8763. PMLR, 2021.

\bibitem{yan2023motion}
Wilson Yan, Andrew Brown, Pieter Abbeel, Rohit Girdhar, and Samaneh Azadi.
\newblock Motion-conditioned image animation for video editing.
\newblock {\em arXiv preprint arXiv:2311.18827}, 2023.

\bibitem{bai2024uniedit}
Jianhong Bai, Tianyu He, Yuchi Wang, Junliang Guo, Haoji Hu, Zuozhu Liu, and Jiang Bian.
\newblock Uniedit: A unified tuning-free framework for video motion and appearance editing.
\newblock {\em arXiv preprint arXiv:2402.13185}, 2024.

\bibitem{xie2023physgaussian}
Tianyi Xie, Zeshun Zong, Yuxin Qiu, Xuan Li, Yutao Feng, Yin Yang, and Chenfanfu Jiang.
\newblock Physgaussian: Physics-integrated 3d gaussians for generative dynamics.
\newblock {\em arXiv preprint arXiv:2311.12198}, 2023.

\end{thebibliography}

%%%%%%%%%%%%%%%%%%%%%%%%%%%%%%%%%%%%%%%%%%%%%%%%%%%%%%%%%%%%

\end{document}